%% file: main.tex
\documentclass[letterpaper]{article} 
\usepackage{aaai25}  
\usepackage{times}  
\usepackage{helvet}  
\usepackage{courier}  
\usepackage[hyphens]{url}  
\usepackage{graphicx} 
\usepackage{natbib}  
\usepackage{caption} 
\usepackage{algorithm}
\usepackage{algorithmic}

\usepackage{multirow}
\usepackage{bm}
\usepackage{xspace}
\usepackage{footnote}
\usepackage{tikz} 
\usepackage{graphicx}
\usepackage{amsmath}
\usepackage{amssymb}
\usepackage{array}
\usepackage{booktabs}

\makeatletter
\DeclareRobustCommand\onedot{\futurelet\@let@token\@onedot}
\def\@onedot{\ifx\@let@token.\else.\null\fi\xspace}

\def\eg{\emph{e.g}\onedot} 
\def\ie{\emph{i.e}\onedot}

\makeatother
\usepackage{epstopdf}
\usepackage{colortbl,xcolor}

\newcommand{\pub}[1]{{\color{gray}{\small{[{#1}]}}}}

\newcommand{\M}{PCAN}
\usepackage{arydshln}

\usepackage{pifont}    
\usepackage{bbding}       

\usepackage{newfloat}
\usepackage{listings}
\DeclareCaptionStyle{ruled}{labelfont=normalfont,labelsep=colon,strut=off} 
\floatstyle{ruled}
\newfloat{listing}{tb}{lst}{}
\floatname{listing}{Listing}
\title{Prototypical Calibrating Ambiguous Samples for Micro-Action Recognition}

\author{
Kun Li\textsuperscript{\rm 1}, Dan Guo\thanks{Corresponding authors}\textsuperscript{\rm 1,2}, Guoliang Chen$^*$\textsuperscript{\rm 1}, Chunxiao Fan\textsuperscript{\rm 1}, Jingyuan Xu\textsuperscript{\rm 1}, Zhiliang Wu\textsuperscript{\rm 3}, \\
Hehe Fan\textsuperscript{\rm 3}, Meng Wang$^*$\textsuperscript{\rm 1,2}}
\affiliations{
    \textsuperscript{\rm 1}School of Computer Science and Information Engineering, Hefei University of Technology \\
    \textsuperscript{\rm 2} Institute of Artificial Intelligence, Hefei Comprehensive National Science Center \\
    \textsuperscript{\rm 3}ReLER Lab, CCAI, Zhejiang University\\

kunli.hfut@gmail.com, guodan@hfut.edu.cn, guoliangchen.hfut@gmail.com, fanchunxiao@hfut.edu.cn, xujingyuan@hfut.edu.cn, wu\_zhiliang@zju.edu.cn, hehefan@zju.edu.cn, eric.mengwang@gmail.com
}

\usepackage{bibentry}

\input{math_commands.tex}

\begin{document}

\maketitle

\begin{abstract}
Micro-Action Recognition (MAR) has gained increasing attention due to its crucial role as a form of non-verbal communication in social interactions, with promising potential for applications in human communication and emotion analysis. However, current approaches often overlook the inherent ambiguity in micro-actions, which arises from the wide category range and subtle visual differences between categories. This oversight hampers the accuracy of micro-action recognition. In this paper, we propose a novel Prototypical Calibrating Ambiguous Network (\textbf{PCAN}) to unleash and mitigate the ambiguity of MAR. \textbf{Firstly}, we employ a hierarchical action-tree to identify the ambiguous sample, categorizing them into distinct sets of ambiguous samples of false negatives and false positives, considering both body- and action-level categories. \textbf{Secondly}, we implement an ambiguous contrastive refinement module to calibrate these ambiguous samples by regulating the distance between ambiguous samples and their corresponding prototypes. This calibration process aims to pull false negative ($\mathbb{FN}$) samples closer to their respective prototypes and push false positive ($\mathbb{FP}$) samples apart from their affiliated prototypes. In addition, we propose a new prototypical diversity amplification loss to strengthen the model's capacity by amplifying the differences between different prototypes. \textbf{Finally}, we propose a prototype-guided rectification to rectify prediction by incorporating the representability of prototypes. Extensive experiments conducted on the benchmark dataset demonstrate the superior performance of our method compared to existing approaches. 
\end{abstract}

\section{Introduction}\label{sec:intro}
Human body actions are an important form of non-verbal communication in social interactions~\cite{aviezer2012body}. In recent years, there has been increasing interest in human-centric action understanding~\cite{balazia2022bodily,liu2021imigue,chen2023smg,guo2024benchmarking,li2023joint,li2024repetitive}. For instance, body language in naturalistic multiview group conversations influences perceptions of leadership and relationships during social interactions~\cite{balazia2022bodily,li2023data}. 
Micro-Action (MA), as a subset, has emerged as a significant research area due to its potential applications in human-to-human communication and analyzing human emotional states~\cite{liu2021imigue,chen2023smg,guo2024benchmarking,guo2024mac,li2024mmad}. 

\begin{figure}[!t]
\centering
\includegraphics[width=1.0\linewidth]{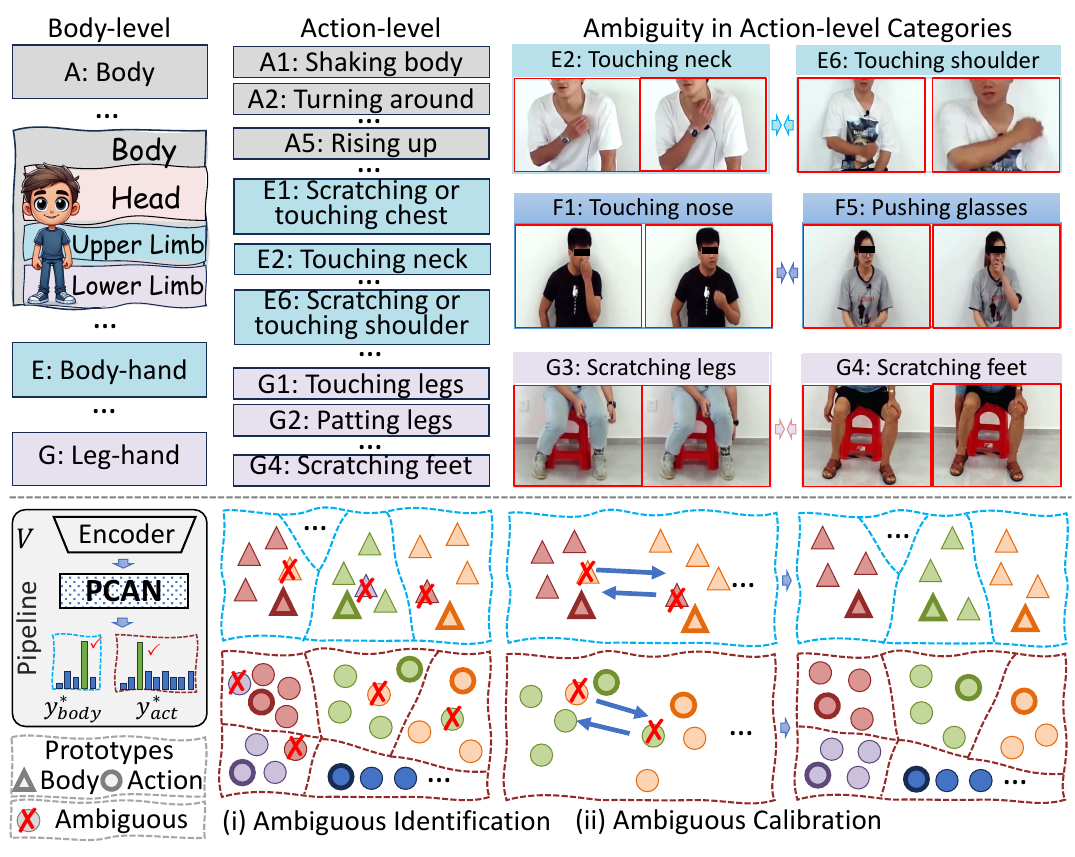}
\caption{
\textbf{TOP}: Micro-Action Recognition (MAR) aims to recognize the body-level and action-level micro-action categories, particularly when dealing with ambiguous samples. 
For example, ``touching shoulder'' and ``touching neck'' belong to the ``body-hand'' but exhibit subtle visual differences. 
\textbf{BOTTOM}: Our approach is motivated by the need to address ambiguities in MAR. 
We begin by identifying ambiguous samples (marked in {\color{red}\ding{55}}) that are prone to misclassification. 
Subsequently, we construct prototypes for each category within the body and action levels, and then align the ambiguous samples with the corresponding prototypes within the feature space. Please Zoom in for details.}
\label{fig:intro}
\end{figure}

As body language is an essential part of understanding human emotions according to psychological studies~\cite{shiffrar2011seeing,aviezer2012body}, iMiGUE~\cite{liu2021imigue} and SMG~\cite{chen2023smg} explored spontaneous micro-gestures in upper limbs driven by a person's inner feelings, revealing the deep emotional states hidden behind the subtle actions. 
To better analyze and understand the spontaneous micro-actions, whole-body movement, and fine-grained action labels are crucial. 
Therefore, \cite{guo2024benchmarking} collected a large-scale spontaneous whole-body micro-actions dataset that includes the head, hands, and upper and lower limbs, using face-to-face psychological interviews way. 
This work extends micro-action from the upper limb to the whole body and builds coarse-to-grained action understanding, benchmarking the datasets, and bringing challenges for human-centric micro-action understanding and analysis. 
Compared to generic action recognition~\cite{lin2019tsm,feichtenhofer2019slowfast,liu2025dynamic,mou2024compressed} and skeleton-based action recognition~\cite{wang20233mformer,zhu2022selective,hao2021hypergraph}, 
micro-actions encompass a multitude of approximate intricate samples, extending across various body regions. 

As shown in Fig.~\ref{fig:intro}, MAR aims to recognize categories at the \textit{body-level} and the \textit{action-level}. \textit{Body-level} category denotes the body part of micro-action occurring, such as body, body-hand, and leg-hand. \textit{Action-level} category denotes the exact name of micro-actions. 
Intuitively, body-level and action-level categories enjoy the nature of hierarchy.   
The actions of ``touching shoulder'' and ``touching neck'' are very similar, both involving hand movement around the upper body. However, current methods typically overlook the inherent ambiguity in micro-actions. 
To reduce the ambiguity in MAR, one solution is to build prototypes that represent the features of each category. However, most of the existing works on few-shot~\cite{zhu2021few,wang2024clip} or egocentric action recognition~\cite{wang2021interactive,dai2023slowfast} and can not be directly applied to tackle the challenges associated with action ambiguity in MAR. 

To address the challenges of ambiguity of micro-actions, we propose a Prototypical Calibrating Ambiguous Network (\M{}). As shown in Fig~\ref{fig:main}, PCAN contains the following designs: 
1) \textbf{Ambiguous Samples Identification} (\S\ref{sec:am_discover}). 
Guided by preliminary predicted scores from the backbone model, we initially identify ambiguous samples at body-level and action-level. 
Concurrently, we mine confidence samples - those correctly classified based on the predicted scores - to capture the fundamental essence and overarching characteristics of action, ensuring the integrity of the prototypes.  
2) \textbf{Ambiguous Samples Contrastive Calibration} (\S\ref{sec:ascc}). 
The identified ambiguous samples are further categorized into false negative ($\mathbb{FN}$) and false positive ($\mathbb{FP}$). 
We then apply contrastive refinement to enhance the representability of prototypes by pulling $\mathbb{FN}$ samples closer to their corresponding prototypes and pushing $\mathbb{FP}$ samples away.
Additionally, we amplify the differences between distinct prototypes to strengthen the model's ability to understand inter-class differences.
3) \textbf{Prototype-guided Rectification} (\S\ref{sec:ppr}). 
Considering that prototypes represent the intrinsic characteristics of each category, we propose a prototype-guided prediction rectification module. By evaluating the similarities between input samples and prototypes, this module refines the model's classification outcomes for MAR. 
Overall, our contributions are summarized as follows. 
\begin{itemize}
\item We attempt to address the ambiguity of micro-actions with prototype learning, wherein prototypes serve as effective representations for discerning micro-actions, providing a compact representation for each category. 

\item We build a prototype-guided rectification approach is implemented to enhance model prediction performance. By evaluating dissimilarities between input samples and prototypes corresponding to different categories, prediction outcomes are rectified. 

\item Extensive experiments conducted on the public micro-action dataset, MA-52, validate the effectiveness of the proposed method. Comparative analysis highlights the effectiveness of the prototypical calibrating ambiguous network in micro-action recognition. 
\end{itemize}

\section{Related Work}\label{sec:related_works}

{\textbf{Micro-Action Recognition}
Micro-Actions~\cite{guo2024benchmarking} refer to subtle and rapid movements that occur across various body parts~\cite{liu2021imigue,chen2023smg,liu2024micro}. 
Existing researches primarily adopt the generic action recognition methods (\eg, 2D CNN based~\cite{lin2019tsm}, 
3D CNN based~\cite{C3D}, 
GCN-based~\cite{yan2018spatial}, and Transformer-based~\cite{liu2022video}) to realize micro-action recognition. 
However, addressing the challenges of subtle differences and high similarity remains difficult with generic action recognition methods. Effectively capturing micro-actions demands in-depth understanding and specialized modeling. 
We conduct experiments on the first human-centered whole-body micro-action dataset named MA-52~\cite{guo2024benchmarking}. It stands out for its extensive range of micro-action categories and abundant samples, encompassing both similar and distinct micro-action categories.

\noindent {\textbf{Prototype Learning for Action Recognition}}
Prototype learning aims to obtain representative feature representation, and has demonstrated substantial potential in addressing the challenges of few-shot scenarios~\cite{zhu2021few,wang2024clip}, egocentric perspectives~\cite{wang2021interactive,dai2023slowfast}, and early action prediction in action recognition~\cite{camporese2023early}. 
However, in the domain of micro-action recognition, action ambiguity remains a significant challenge due to the subtle differences between actions, which can lead to misclassification. 
Previous prototype learning methods can partially mitigate this issue by creating more discriminative and representative prototypes. Despite these advancements, they do not fully resolve the challenges associated with action ambiguity in micro-actions. 
In this work, we separate the ambiguous samples into sets of false positives and false negatives with the guidance of prototype learning and calibrate these ambiguous samples according to the prototypical distance measurement with their affiliated prototypes. 

\begin{figure*}[!t]
\centering
\includegraphics[width=1.0\linewidth]{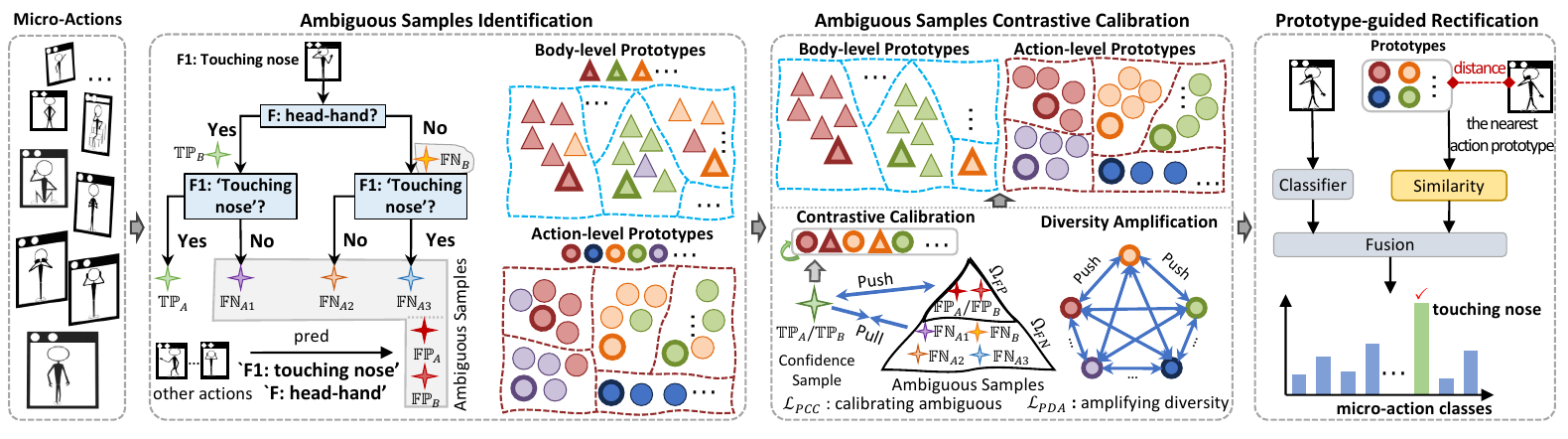}
\caption{The overview of Prototypical Calibrating Ambiguous Network (\M{}). 
In \textbf{Ambiguous Samples Identification} (\S\ref{sec:asi}), we discover ambiguous samples ($\mathbb{FN}$ and $\mathbb{FP}$) through the preliminary prediction scores. 
In \textbf{Ambiguous Samples Contrastive Calibration} (\S\ref{sec:ascc}), we use contrastive prototype calibration and prototype diversity amplification losses to calibrate the prototypes and eliminate the influence of ambiguous samples. 
In \textbf{Prototype-guided Rectification} (\S\ref{sec:ppr}), we incorporate the similarity between established prototypes and video embedding for action-level category prediction. 
}
\label{fig:main}
\end{figure*}

\noindent {\textbf{Ambiguous/Hard Sample Learning}}
The ambiguous samples are difficult to classify or distinguish clearly due to factors such as similar actions, occlusion, variety, or data quality issues~\cite{xia2023towards}. 
In recent years, effectively distinguishing ambiguous samples has proven a difficult challenge~\cite{li2020detailed,li2021spherical,zhou2023learning}. 
For example, \cite{zhou2023learning} utilized prototypical contrastive learning to constrain the distance between confident samples and ambiguous samples to improve the recognition of ambiguous actions that are quite similar and easily misclassified. 
However, all of the above methods only consider prototypes at a certain grain and lack consideration of cascade coarse- and fine-grained label categories. 
We propose cascade prototypes that work to discover and calibrate ambiguous samples between these micro-actions in order to reduce the category ambiguity between them.

\section{Methodology}
\label{sec:method}

\subsection{Preliminary and Core Idea}
\subsubsection{Preliminary.} 
Give a video $V$, the \emph{Micro-Action Recognition} (MAR) task~\cite{guo2024benchmarking} predicts the micro-action categories at {body-level} $\gY_{body}\in\{1,2,\ldots,N_{B}\}$ and at {action-level} $\gY_{act}\in\{1, 2, \ldots, N_{A}\}$, where $N_{B}$ and $N_{A}$ denotes the number of body and action categories, respectively. 
PoseConv3D~\cite{duan2022revisiting} is a novel framework for RGB-Pose dual-modality action recognition. 
Concretely, the input RGB data $\mathcal{D}_{RGB}$ and skeleton data $\mathcal{D}_{Pose}$ are fed to separate 3D-CNN based encoder $\mathcal{E}_{R}$ and $\mathcal{E}_{P}$ to extract the feature. 
After that, the predictions of the RGB/Pose are combined through late fusion to compute class scores for action recognition. In this work, we conduct experiments on PoseConv3D to validate the motivation to remove ambiguity in MAR. 
Considering that RGB and Pose are parallel operations, we take the RGB branch as an example to elaborate. 

\subsubsection{Core Idea.} 
In this study, we argue that the misclassification in MAR primarily arises from the inherent ambiguity in subtle differences among action-level categories. 
To address this, we first identify the ambiguous samples at the body and action levels, which are based on the preliminary classification results. 
Simultaneously, we mine the confidence samples to build body- and action-level prototypes. 
Then, we utilize a hierarchical contrastive calibration mechanism to reduce ambiguity by regulating the distance between ambiguous samples and their corresponding prototypes. 
Finally, we leverage the representability of the prototype to perform prototype-guided rectification for action-level classification. 

\subsection{Ambiguous Samples Identification}\label{sec:asi}
\subsubsection{Action-tree Hierarchy Modeling.} \label{sec:at}
In micro-action recognition, the categories at the body level and action level naturally follow a hierarchical structure. 
Namely, the Micro-Action existing ``\textbf{body}$\prec$\textbf{action}'' structural feature, namely, each \textbf{action-level} category belongs to a specific primary \textbf{body-level} category. 
For instance, MA ``B1: nodding'' has a corresponding body label ``B: head''. 
Inspired by this, we propose a hierarchical action-tree structure to model the relation between body- and action-level categories. 
Let $F_i\in\R^d$ be the learned feature from the backbone of a sample $i$ and two classifiers $\text{CLS}_B(\cdot)$ and $\text{CLS}_A(\cdot)$ to predict the probability of action-level and body-level category of $F_i$. 
To inject the hierarchy structure into action prediction, we build the hierarchical action probability $q_{J}$: 
\begin{equation}
q_{J} = \lambda \cdot \text{CLS}_{B}(F_i) \oplus\text{CLS}_{A}(F_i) = \lambda \cdot q_{B} \oplus q_{A},
\label{eq:p}
\end{equation}
where $\lambda$ is a hyper-parameter that is optimally set to 1 and discussed in \S\ref{sec:abl}, and $\oplus$ is the operation to incorporate the body- and action-level probabilities. $q_{B}$ and $q_{A}$ denote the body-level and action-level probability, respectively. 
To optimize $q_{J}$, we define a hierarchical probabilistic loss $\mathcal{L}_{HP}$ to maximize the marginal probability of action-grained predictions by aggregating the structural information of the relevant labels defined as described in the action-tree:
\begin{equation}
\mathcal{L}_{HP}=- \log {\frac{\exp(q_J^{GT})}{\sum_{i=1}^{N_B+N_A}\exp(q_J^i)}},
\end{equation}
where $q_J^{GT}$ is the probability score corresponding to the index of ground truth of the action-level label. Note that $q_{J}$ is only used in loss $\mathcal{L}_{HP}$. 

\subsubsection{Ambiguous Samples.} 
As stated in \S\ref{sec:intro}, the ambiguity of micro-actions is caused by the minor differences in action-level, which will hinder the model from achieving optimal results. 
Based on the classification results from classifiers $\mathrm{CLS}_B$ and $\mathrm{CLS}_A$, we can identify the ambiguous samples. 
\textbf{(i) Ambiguous Samples - $\mathbb{FN}$ set.} 
In the classification task, $\mathbb{FN}$ (False Negative) refers to instances where the model incorrectly predicts the negative class when the true class is positive.  
If we merely consider the body-grained, instances of body misclassified as other categories are termed as false negatives ($\mathbb{FN}_B$). 
We also can consider both body- and action-level predictions. If body-level is correct but action-level is wrong, it is denoted as $\mathbb{FN}_{A1}$; if both body-level and action-level are wrong, it is denoted as $\mathbb{FN}_{A2}$; if body-level is wrong but action-level is correct, it is denoted as $\mathbb{FN}_{A3}$. 
\textbf{(ii) Ambiguous Samples - $\mathbb{FP}$ set.}
In action-level, if samples from other categories are mistakenly classified as the current category, they are labeled as $\mathbb{FP}_A$ samples. 
This split is evaluated solely at the action-grained level. 
Instances from other categories misclassified as the referred body-grained category are referred to as false positives ($\mathbb{FP}_B$). 
Please note that the identification of ambiguous samples occurs online during training, and we built ambiguous sample sets for both body-level and action-level categories.

\subsubsection{Hierarchical Prototype Initialization.} 
Considering the two-level category in MAR, we define the body-level prototypes as $\mP_{B}=\{p_1, p_2, \ldots, p_{N_B}\}$, and the action-level prototypes as $\mP_{A}=\{p_1, p_2, \ldots, p_{N_A}\}$. 
$\mP_{B}\in\R^{N_B\times d}$ and $\mP_{A}\in\R^{N_A\times d}$ serve as the stable estimate of the clustering center for each micro-action category. Initially, these prototypes are randomly initialized and then optimized throughout the training process. 
In prototype learning, correctly predicted samples capture the fundamental essence and overarching characteristics of an action. These samples are considered confident samples. 
The body-level confident sample set is defined as $\mathbb{TP}_B$ for samples that correctly predicted their body-level labels. 
Considering the ambiguity of action-level, the samples predicted at both body-level and action-level are called confident sample sets, denoted as $\mathbb{TP}_A$. 
Thus, we have $\Omega_{TP}=\{\mathbb{TP}_B,\mathbb{TP}_A\}$, which represents two kinds of confident samples in the body- and action-level prototypes. 

\subsubsection{Online Prototype Update.}\label{sec:am_discover}
Since the confidence samples ($\Omega_{TP}$) usually have better intra-class consistency, we use it to update the prototype representation of the corresponding category $k$ by exponential moving average (EMA)~\cite{ye2019unsupervised,wu2018unsupervised,wang2022renovate}: 
\begin{equation} 
p_{k}=(1- \rho)\cdot \frac{1}{N^k_{TP}} \sum\limits_{i \in \Omega^k_{TP}}F_i+\rho \cdot p_{k}^{pre},
\end{equation}
where $F_i$ is the feature of $i$-th sample, $p_{k}^{pre}$ is the prototype before updating, $N_{TP}^k$ is the size of $\mathbb{TP}$ sample set $\Omega_{TP}$ under category $k$, and $\rho$ is a momentum term and empirically set to 0.9 as in~\cite{zhou2023learning}.

\subsection{Ambiguous Samples Contrastive Calibration}\label{sec:ascc}
Through the above operations, we obtained ambiguous samples and prototypes. 
However, the ambiguous samples and prototypes are misaligned in the feature space due to the ambiguity of micro-actions. 
To address this issue, we utilize a contrastive prototype loss~\cite{zhou2023learning,wang2022renovate} to regulate the distance between ambiguous samples and their prototypes.
Concretely, we aim for $\mathbb{FP}$ samples to be distant from the misassigned prototype and for $\mathbb{FN}$ samples to be close to the correct prototype. 
Here, we use the action-level prototype and ambiguous samples as examples to detail the contrastive calibration. The approach of action-level calibration is analogous. 

\subsubsection{Prototypical Contrastive Calibration} 
For the ambiguous sample sets $\Omega_{FN}$ ($\mathbb{FN}$ samples) and $\Omega_{FP}$ ($\mathbb{FP}$ samples), we first estimate their mean values as the center representations in the feature space: 
\begin{equation}
\mu_{\omega}^k= \frac{1}{N_{\omega}^k} \sum\limits_{i \in \Omega^{k,\omega}_{FN}}F_i, \quad \  
\mu_{FP}^k= \frac{1}{N^k_{FP}} \sum\limits_{i \in \Omega^k_{FP}}F_i,
\end{equation}
where $k$ is action category, ${\omega}$ denotes the type of ambiguous samples in the $\mathbb{FN}$ set $\Omega_{FN}=\{\mathbb{FN}_{A1}, \mathbb{FN}_{A2}, \mathbb{FN}_{A3}\}$ defined in \S\ref{sec:am_discover}. 
$N_{\omega}^k$ and $N_{FP}^k$ are the sizes of sample sets $\Omega^{k,\omega}_{FN}$ and $\Omega^k_{FP}$ under category $k$, respectively.

\textbf{(i) Ambiguous Samples - $\mathbb{FN}$ set.} 
The auxiliary term $\psi_i$ is used to calibrate the model's predictions for true action categories as,  
\begin{equation}
\psi_i^\omega = 
\begin{cases} 
1 - dis(F_i, \mu_\omega^k), & \text{if } N_\omega^k > 0 \text{ and } i \in \Omega_{FN}^{k,\omega}; \\
0, & \text{otherwise},
\end{cases}
\label{eq:psi}
\end{equation}
where $dis(,)$ denotes the cosine distance between two features. 
This operation brings the ambiguous samples closer to the confidence samples (\ie, prototypes) in the feature space. 
This adjustment helps rectify the predicted logits of the ambiguous samples, aligning them more closely with the correct action categories. By minimizing $\psi_i^\omega$, the model more accurately identifies the ambiguous samples, as a smaller value indicates that the ambiguous samples are in closer proximity to the prototypes. 
\textbf{(ii) Ambiguous Samples - $\mathbb{FP}$ set.} 
For ambiguous samples $\mathbb{FP}$ in set $\Omega_{FP}$, we define the auxiliary term $\Psi_i$, which serves to adjust the predictions for non-true action categories as, 
\begin{equation}
\Psi_i=\left\{\begin{matrix} 
1+dis(F_i,\mathbf{\mu}_{FP}^{k}), \text{if}\ N_{FP}^{k}>0 \ \text{and} \ i \in \Omega^k_{FP};  \\  
0, \ \text{otherwise},
\end{matrix}\right. 
\label{eq:Psi}
\end{equation}
where $\Psi_i^\omega$ is the penalty term for sample sets in $\Omega_{FP}$, which can keep the ambiguous samples in $\Omega_{FP}$ away from the confidence samples in the feature space. 
By minimizing $\Psi_i^\omega$, the distance between ambiguous samples and prototypes can be increased, thus mitigating the risk of the model incorrectly classifying ambiguous samples as action category $k$. 

In order to keep false positive ($\mathbb{FP}$ in $\Omega_{FP}$) samples distant from prototype and false negative ($\mathbb{FN}$ in $\Omega_{FN}$) samples close to the prototype, we define a matrix $\mathbf{Z}_i^{\omega}$ to represent the distance between the ambiguous samples and their prototype as follows:
\begin{equation}
\left\{
\begin{aligned}
\mathbf{Z}_i^{\omega} &= \tau \cdot dis(F_i, p_k) - (1 - q_{ik}) \psi_i^{\omega}, && \text{if } i \in \Omega_{FN}^{\omega}; \\
\mathbf{Z}_i^{FP} &= \tau \cdot dis(F_i, p_k) - (1 - q_{ik}) \Psi_i, && \text{if } i \in \Omega_{FP},
\end{aligned}
\right.
\label{eq:y}
\end{equation}
where $p_k$ denotes the prototype feature of action $k$, $\tau$ is a hyper-parameter {and is set to 0.125}, $q_{ik}$ is the probability of sample $i$ in micro-action $k$. 
Thus, the prototype contrastive calibration loss $\mathcal{L}_{PCC}$ is formulated as follows:
\begin{equation}
\begin{aligned}
\mathcal{L}_{PCC} = & - \sum_{i \in \Omega_{FN}^{\omega}} \log \frac{e^{\mathbf{Z}_{i}^{\omega}}}{e^{\mathbf{Z}_{i}^{\omega}} + \sum_{l \neq k} e^{dis(F_i, p_l) \cdot \tau}}   \\
& - \sum_{j \in \Omega_{FP}} \log  \frac{e^{\mathbf{Z}_{i}^{FP}}}{e^{\mathbf{Z}_{i}^{FP}} + \sum_{j \neq k} e^{dis(F_i, p_j) \cdot \tau}},
\end{aligned}
\label{eq:intra}
\end{equation}
where ${\omega} \in \{\mathbb{FN}_{A1}, \mathbb{FN}_{A2}, \mathbb{FN}_{A3}\}$ denotes the type of ambiguous samples in $\Omega_{FN}$ defined in Sec.~\ref{sec:am_discover}. 
We employ $\alpha_1$, $\alpha_2$, and $\alpha_3$ as the weights to balance the set of $\Omega$, and discussed in \S\ref{sec:abl}. 

\subsubsection{Prototypical Diversity Amplification}
\label{sec:pd}
As stated in \S\ref{sec:intro}, the ambiguity is caused by the high similarity of action-level categories under the same body part. 
For instance, head movements like ``head up'' and ``nodding'' enjoy similar motion patterns, leading to difficulty in differentiation. 
To mitigate the influence of high similarity among different categories, a prototypical diversity amplification is designed to amplify the diversity among different categories: 
\begin{equation}
\mathcal{L}_{PDA}= \sqrt{ \sum\limits_{j=1}^{N_A} \sum\limits_{i=1}^{N_A}  \frac{ p_i \cdot p_j }{ \parallel p_i\parallel \parallel p_j\parallel}},
\end{equation}
where $p_i$ and $p_j$ stand for action-level prototypes. 

\subsection{Prototype-guided Rectification}
\label{sec:ppr}
In the previous section, we established prototypes for each category, providing a clear reference that represents the intrinsic characteristics of different categories. 
Building on this, we propose a prototype-guided rectification module that rectifies prediction results by analyzing the cosine similarity between input samples and established prototypes. 
By comparing the cosine similarity between prototypes and output features, we can capture subtle differences missed in the initial prediction. This adjustment refines the final prediction, improving both accuracy and reliability. 
The final prediction probability of sample $F_i$ is as follows:
\begin{equation}
F_{act, i}=\mW_1 F_{i} + b_1 +\gamma\cdot dis(\mW_2 F_{i} +b_2, \mP_{A}),
\label{eq:gamma_add}
\end{equation}
\begin{equation}
y_{act, i}^{\ast}=\mathop{\arg\max}\limits_{y\in \gY_{act}}(\texttt{SoftMax}(F_{act, i})),
\end{equation}
where $\mW_1\in\R^{d\times N_A}$ and $\mW_2\in\R^{d\times N_A}$ are two learnable matrices to transform the embedding $F_i$. $b_1$ and $b_2$ are two biases. $\mP_{A}$ is the prototypes of action-level categories. $\gamma$ serves as a balancing weight. 
For body-level probability $y_{body, i}^{\ast}$, we adopt a body-level classifier to predict it. 

\subsection{Model Optimization}
\label{sec:optimi}
We use the cross-entropy loss as the basic objective for the micro-action recognition task, denoted as $\mathcal{L}_{CE}$. 
Considering that \M{} is RGB-Pose dual-modality based, we implement losses $\mathcal{L}_{CE}$, $\mathcal{L}_{HP}$, $\mathcal{L}_{PCC}$ and $\mathcal{L}_{PDA}$ in both RGB and Pose. 
Finally, the total loss is formulated as follows:
\begin{equation}
\mathcal{L}=\mathcal{L}_{CE}+ \mathcal{L}_{HP} + \mathcal{L}_{PCC}+\beta \mathcal{L}_{PDA},
\label{eq:loss}
\end{equation}
where $\beta$ is the balance hyperparameters for the loss term. 

\section{Experiments}
\label{sec:experiment}
\subsection{Experimental Setup}
\noindent\textbf{Micro-Action Dataset.} 
{MA-52}~\cite{guo2024benchmarking} 
\footnote{\url{https://github.com/VUT-HFUT/Micro-Action}} is a large-scale human-centered whole-body micro-action dataset that takes an interview format to record unconscious human micro-action behaviors. It contains micro-actions in the whole body, including the body, head, upper, and lower limbs. The dataset contains \textit{22,422 (22.4K) videos} interviewed from 205 participants with annotations categorized into two levels: \textit{7 body-level and 52 action-level micro-action categories}, making it a large-scale dataset of the largest samples and micro-action categories that we are aware of to date. The dataset consists of 11,250, 5,586, and 5,586 instances in training/validation/test, respectively. 

\noindent\textbf{Evaluation Metrics.} 
Following the practice~\cite{guo2024benchmarking,guo2024mac}, we adopt the metrics of Top-1/-5 accuracy and macro and micro F1 scores of body and action levels to evaluate the performance of the proposed model.  
F1$_{mean}$ is used as a general evaluation metric, and its formula is as follows:
\begin{equation}
\rm F1_{\emph{mean}} =(\rm F1_{\emph{macro}}^{\emph{body}} \!+\! \rm F1_{\emph{micro}}^{\emph{body}} \!+\! \rm F1_{\emph{macro}}^{\emph{action}} \!+\! \rm F1_{\emph{micro}}^{\emph{action}}) / 4.
\label{eq:map}
\end{equation}

\noindent\textbf{Implementation Details.} 
For each video, each frame is resized into $224\times 224$. 
For model training, we adopt the SGD optimizer with a learning rate of 0.0075, a momentum of 0.9, a weight decay of 1e-4, and a batch size of 10. 
The learning rate is reduced by a factor of 10 at the 15th and 30th epochs, and the model is trained with 40 epochs. 
In Eq.~\ref{eq:gamma_add}, we set $\gamma=1$ for the RGB branch and $\gamma=5$ for the Pose branch. 
In Eq.~\ref{eq:loss}, we set $\beta=5$ for the RGB branch and $\beta=5$ for the Pose branch.

\begin{table}[t!]
\centering
\tabcolsep 2pt
\resizebox{1.0\linewidth}{!}{
\begin{tabular}{l|c|cc|cc|cc|c}
\hline
\multirow{2}{*}{Method} & \textit{Body} & \multicolumn{2}{c|}{\textit{Action}} 
 & \multicolumn{2}{c|}{\textit{Body}} & \multicolumn{2}{c|}{\textit{Action}} & \textit{All} \\ 
& Top-1   & Top-1      & Top-5 & F1$_{macro}$          & F1$_{micro}$         & F1$_{macro}$          & F1$_{micro}$          & F1$_{mean}$         \\ \hline
TSN\hspace{0.2em}\pub{ECCV'16} & 59.22 & 34.46 & 73.34 & 52.50 & 59.22 & 28.52 & 34.46 & 43.67 \\
TIN\hspace{0.2em}\pub{AAAI'20} & 73.26 & 52.81 & 85.37 & 66.99 & 73.26 & 39.82 & 52.81 & 58.22 \\
TSM\hspace{0.2em}\pub{ICCV'19} & 77.64 & 56.75 & 87.47 & 70.98 & 77.64 & 40.19 & 56.75 & 61.39 \\
MANet\hspace{0.2em}\pub{TCSVT'24} & 78.95 & 61.33 & 88.83 & 72.87 & 78.95 & 49.22 & 61.33 & 65.59 \\ \hline
C3D\hspace{0.2em}\pub{ICCV'15} & 74.04 & 52.22 & 86.97 & 66.60 & 74.04 & 40.86 & 52.22 & 58.43 \\
I3D\hspace{0.2em}\pub{CVPR'17} & 78.16 & 57.07 & 88.67 & 71.56 & 78.16 & 39.84 & 57.07 & 61.66 \\
SlowFast\hspace{0.2em}\pub{ICCV'19} & 77.18 & 59.60 & 88.54 & 70.61 & 77.18 & 44.96 & 59.60 & 63.09 \\ \hline
VSwinT\hspace{0.2em}\pub{CVPR'22} & 77.95 & 57.23 & 87.99 & 71.25 & 77.95 & 38.53 & 57.23 & 61.24 \\
TimesFormer\hspace{0.2em}\pub{ICML'21} & 69.17 & 40.67 & 82.67 & 61.90 & 69.17 & 34.38 & 40.67 & 51.53 \\
Uniformer\hspace{0.2em}\pub{TPAMI'23} & 79.03 & 58.89 & 87.29 & 71.80 & 79.03 & 48.01 & 58.89 & 64.43 \\ \hline
STGCN\hspace{0.2em}\pub{AAAI'18} & 69.87 & 49.61 & 79.54 & 61.53 & 69.87 & 34.64 & 49.61 & 53.91 \\
2s-AGCN\hspace{0.2em}\pub{CVPR'19} & 70.07 & 49.48 & 78.27 & 61.30 & 70.07 & 34.64 & 49.48 & 53.87 \\
CTRGCN\hspace{0.2em}\pub{ICCV'21} & 72.06 & 52.61 & 81.22 & 63.46 & 72.06 & 37.79 & 52.61 & 56.48 \\ \hline
B2C-AFM\hspace{0.2em}\pub{TIP'23} & 78.34 & 57.12 & 88.45 & 72.53 & 78.34 & 44.12 & 57.12 & 63.03 \\
PoseConv3D (RGB) & 77.77 & 56.30 & 86.11 & 69.87 & 77.77 & 35.80 & 56.30 & 59.93 \\
PoseConv3D (Pose) & 74.60 & 50.93 & 84.98 & 67.31 & 74.60 & 31.07 & 50.93 & 55.98 \\
PoseConv3D\hspace{0.2em}\pub{CVPR'22} & 80.95 & 63.52 & 90.23 & 74.96 & 80.95 & 47.20 & 63.52 & 66.66 \\ \hline
\rowcolor{gray!20}\textbf{Ours (RGB)} & 79.36 & 60.03 & 87.65 & 72.37 & 79.36 & 43.29 & 60.03 & 63.76 \\
\rowcolor{gray!20}\textbf{Ours (Pose)} & 76.99 & 56.51 & 87.13 & 71.33 & 76.99 & 42.28 & 56.51 & 61.78 \\
\rowcolor{gray!20}\textbf{Ours} & \textbf{82.30} & \textbf{66.74} & \textbf{91.75} & \textbf{77.02} & \textbf{82.30} & \textbf{53.83} & \textbf{66.74} & \textbf{69.97} \\
\hline
\end{tabular}
}
\caption{Performance comparison of micro-action recognition on the MA-52 dataset. Top-1/5 are Top-1/5 accuracy. 
}
\label{tab:comp}
\end{table}

\subsection{Comparison with the State-Of-The-Art}
As depicted in Tab.~\ref{tab:comp}, we compare our method with classical methods in 5 groups, namely, 
\textbf{1) 2D CNN based} (\ie, TSN~\shortcite{wang2016temporal}, TIN~\shortcite{shao2020temporal}, TSM~\shortcite{lin2019tsm} and MANet~\shortcite{guo2024benchmarking}),  
\textbf{2) 3D CNN based} (\ie, C3D~\shortcite{C3D}, I3D~\shortcite{I3D} and SlowFast~\shortcite{feichtenhofer2019slowfast}), \textbf{3) Transformer-based} (\ie, VSwinT~\shortcite{liu2022video}, TimesFormer~\shortcite{bertasius2021space} and Uniformer~\shortcite{li2023uniformer}), \textbf{4) GCN based} (\sloppy STGCN~\shortcite{yan2018spatial}, 2s-AGCN~\shortcite{shi2019two} and CTRGCN~\shortcite{chen2021channel}), and \textbf{5) Dual-modality based} (B2C-AFM~\shortcite{guo2023b2c} and PoseConv3D~\shortcite{duan2022revisiting}). 
In 2D CNN based methods, MANet achieves 65.59\% on $\rm F1_{mean}$, while the CTRGCN~\cite{chen2021channel} in GCN-based methods only achieves 56.48\%, which is 9.11\% lower than MANet. This is due to the fact that skeleton information is more susceptible to the quality of key points obtained by the pose estimator, and micro-actions usually have short durations and low action amplitudes, which leads to the difficulty in improving the performance of the GCN-based methods. 
On the action-level $\rm F1_{micro}$ metric, PoseConv3D (RGB) and PoseConv3D (Pose) achieve 56.30\% and 50.93\%, and the performance can reach 63.52\% with two-modality fusion. Whereas our method achieves 60.03\% and 56.51\% for RGB and Pose branches, the performance can be improved to 66.74\% after fusion. This proves that the proposed prototype calibration method can be effective for multiple modalities.
The $\rm F1_{macro}$ in the action-level categories of our method reaches 53.83\%, which is 6.63\% higher than PoseConv3D, indicating that we are able to distinguish the ambiguous samples effectively. 
It proves the proposed method can improve the recognition accuracy in micro-action across different action-level categories.

\subsection{Ablation Studies}\label{sec:abl}

\begin{table}[t!]
\centering
\tabcolsep 4pt
\resizebox{1.0\linewidth}{!}{
\begin{tabular}{c|c|cc|cc|cc|c}
\hline
\multirow{2}{*}{$\lambda$} & \textit{Body} & \multicolumn{2}{c|}{\textit{Action}} & \multicolumn{2}{c|}{\textit{Body}} & \multicolumn{2}{c|}{\textit{Action}} & \textit{All}  \\ 
& Top-1   & Top-1      & Top-5 & F1$_{macro}$          & F1$_{micro}$         & F1$_{macro}$          & F1$_{micro}$          & F1$_{mean}$         \\ \hline
0 &81.76 &66.34 &\textbf{91.80} &76.26 &81.76 &52.69 &66.34 &69.26 \\
0.1 & 81.99 &66.27 &91.14 &76.65 &81.99 &52.28 &66.27 &69.30 \\
10 &81.95 &66.29 &91.48 &76.69 &81.95 &51.82 &66.29 &69.19 \\
\rowcolor{gray!20} 1 &\textbf{82.30} &\textbf{66.74} &91.75 &\textbf{77.02} &\textbf{82.30} &\textbf{53.83} &\textbf{66.74} &\textbf{69.97} \\
 \hline
\end{tabular}
}
\caption{Ablation for $\lambda$ in $\mathcal{L}_{HP}$ on the MA-52 dataset. }
\label{tab:abl-actiontree}
\end{table}

\begin{table}[t!]
\centering
\footnotesize
\tabcolsep 2pt
\resizebox{1.0\linewidth}{!}{
\begin{tabular}{ccc|c|cc|cc|cc|c}
\hline
\multirow{2}{*}{$\alpha_1$} & \multirow{2}{*}{$\alpha_2$} &\multirow{2}{*}{$\alpha_3$} & \textit{Body} & \multicolumn{2}{c|}{\textit{Action}}  & \multicolumn{2}{c|}{\textit{Body}} & \multicolumn{2}{c|}{\textit{Action}} & \textit{All} \\ 
& & & Top-1   & Top-1      & Top-5  & F1$_{macro}$          & F1$_{micro}$         & F1$_{macro}$          & F1$_{micro}$          & F1$_{mean}$       \\ \hline
1 & 1  & 1  &82.04 &66.41 &91.64 &76.59 &82.04 &53.44 &66.61 &69.67  \\ 
1 &0.1 &0.1 &81.90 &66.47 &91.51 &76.09 &81.90 &52.79 &66.47 &69.31 \\
0.1 &1 &0.5 &81.92 &66.13 &91.60 &76.31 &81.92 &52.49 &66.13 &69.21 \\ 
1 &0.5 &1   &81.83 &66.06 &91.34 &75.89 &81.83 &52.23 &66.06 &69.00 \\
1 &0.1 &0.5 &82.01 &66.67 &91.48 &76.32 &82.01 &53.00 &66.67 &69.50     \\
\rowcolor{gray!20}1 &0.5 &0.1 &\textbf{82.30} &\textbf{66.74} &\textbf{91.75} &\textbf{77.02} &\textbf{82.30} &\textbf{53.83} &\textbf{66.74} &\textbf{69.97} \\
\hline
\end{tabular}
}
\caption{
Ablation results for different weights of ambiguous sample sets on the MA-52 dataset. $\alpha_1$, $\alpha_2$ and $\alpha_3$ are weights for $\mathbb{FN}_{A1}$, $\mathbb{FN}_{A2}$ and $\mathbb{FN}_{A3}$, respectively. 
}
\label{tab:fn123}
\end{table}

\textbf{Action-tree Hierarchy Modeling.} 
The body-level micro-actions are represented as parent nodes in the action-tree, while the action-level micro-actions are depicted as child nodes positioned under their corresponding parent nodes, which models the hierarchical structural relationship. In Tab.~\ref{tab:abl-actiontree}, we conduct ablation experiments with the hyper-parameter $\lambda$ in Eq.~\ref{eq:p} on the RGB branch. When $\lambda=0$, it signifies that the action-level probabilities are not influenced by the action-tree structure. Regarding the $\rm F1_{macro}$ score for action-level, $\lambda=1$ yields a 1.14\% improvement compared to $\lambda=0$. 
This proves that the proposed action-tree structure effectively improves the accuracy of identifying action-level micro-actions.

\noindent\textbf{Ambiguous Samples Contrastive Calibration.} 
Here, we explore the impact of hyper-parameters on ambiguous sample sets $\mathbb{FN}_{A1}$, $\mathbb{FN}_{A2}$, and $\mathbb{FN}_{A3}$ as stated in prototype identification. 
Each of these sets represents distinct types of ambiguous samples. 
For $\mathbb{FN}_{A1}$, the model correctly captures most of the general features of the sample, but makes errors in the action-level category. 
Hence, we assign a higher weight to emphasize the model's inaccuracies in the finer details. 
For $\mathbb{FN}_{A2}$, while the model fails at both scales, errors at the higher are more critical than those at $\mathbb{FN}_{A1}$. 
For $\mathbb{FN}_{A3}$, these ambiguous samples result from the model's misreading of the overall characteristics, so we assign them lower weights.  
We experiment with various weight configurations to investigate the importance of different types of ambiguous sample sets, as shown in Tab.~\ref{tab:fn123}. 
The optimal result is achieved when weights for $\mathbb{FN}_{A1}$, $\mathbb{FN}_{A2}$, and $\mathbb{FN}_{A3}$ decreased step by step. 
This indicates that the $\mathbb{FN}_{A1}$ set significantly contributes to the calibration of ambiguous samples. Despite the model accurately classifying overall actions, it struggles with similar micro-action categories, necessitating further calibration. 
Additionally, when body-level labels are incorrectly predicted, including $\mathbb{FN}_{A2}$ and $\mathbb{FN}_{A3}$ samples, they are often attributed to factors like data noise or sample imbalance. 
Compared to the $\mathbb{FN}_{A1}$ set, these samples are less reliable and of lower quality. Therefore, they require further calibration with a separate cluster center and a relatively low-weight assignment.

\begin{figure}[t!]
\centering
\includegraphics[width=1.0\linewidth]{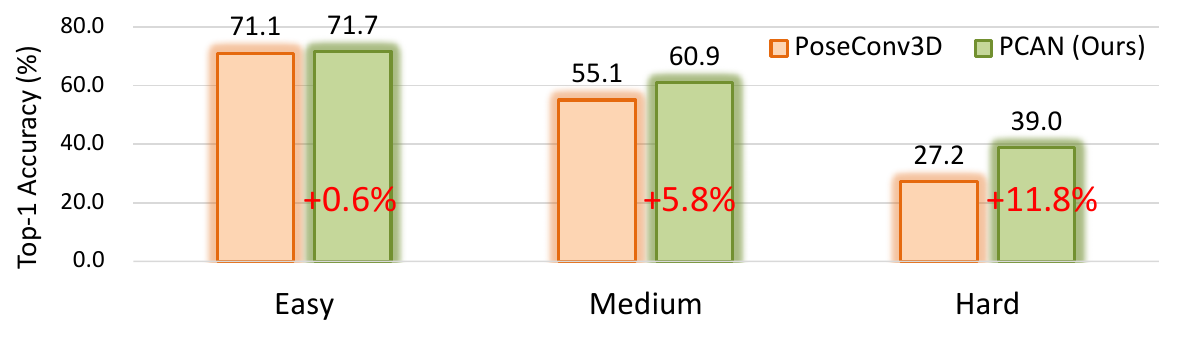}
\caption{Top-1 Accuracy (\%) on ambiguous micro-actions for the MA-52 dataset.}
\label{fig:easy_hard}
\end{figure}

\begin{figure*}[t!]
\centering
\includegraphics[width=1.0\linewidth]{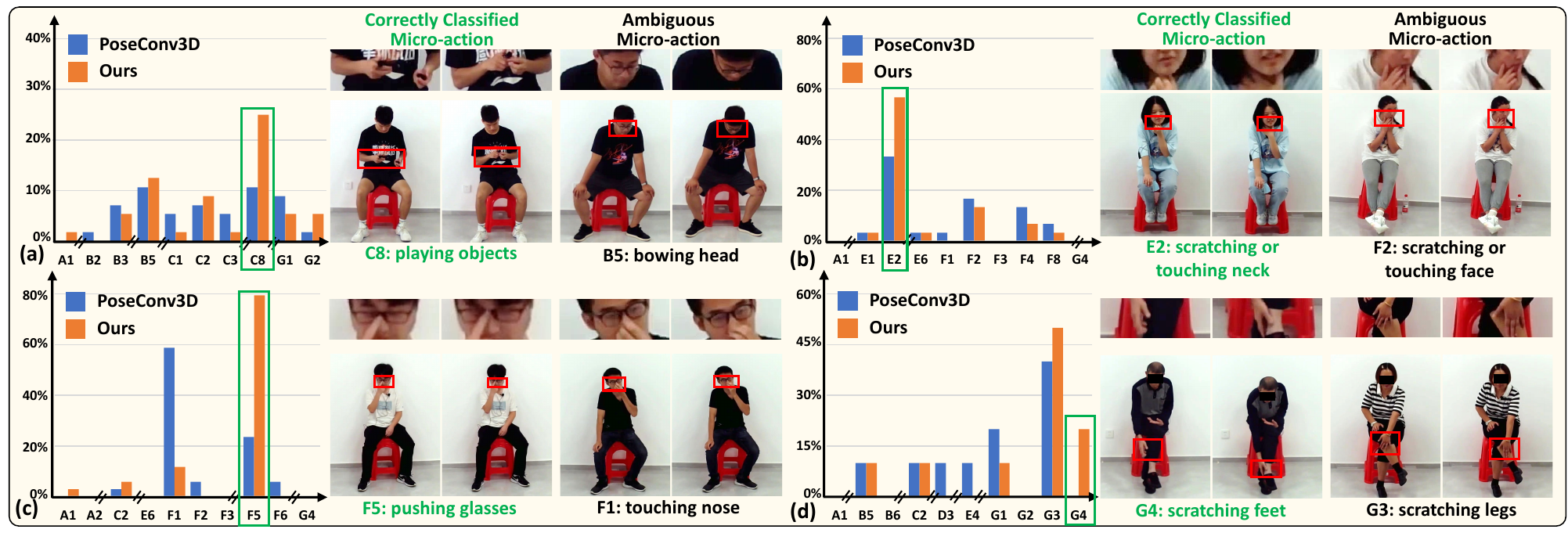}
\caption{Comparative analysis of the PoseConv3D~\cite{duan2022revisiting} and ours \M{}. 
The green categories are the micro-action categories being explored. Our \M{} method demonstrates an excellent ability to categorize ambiguous action categories. 
The proposed \M{} exhibits robust performance at both body-level and action-level. 
}
\label{fig:app_vis3}
\end{figure*}

\begin{figure*}[!t]
\centering
\includegraphics[width=1.0\linewidth]{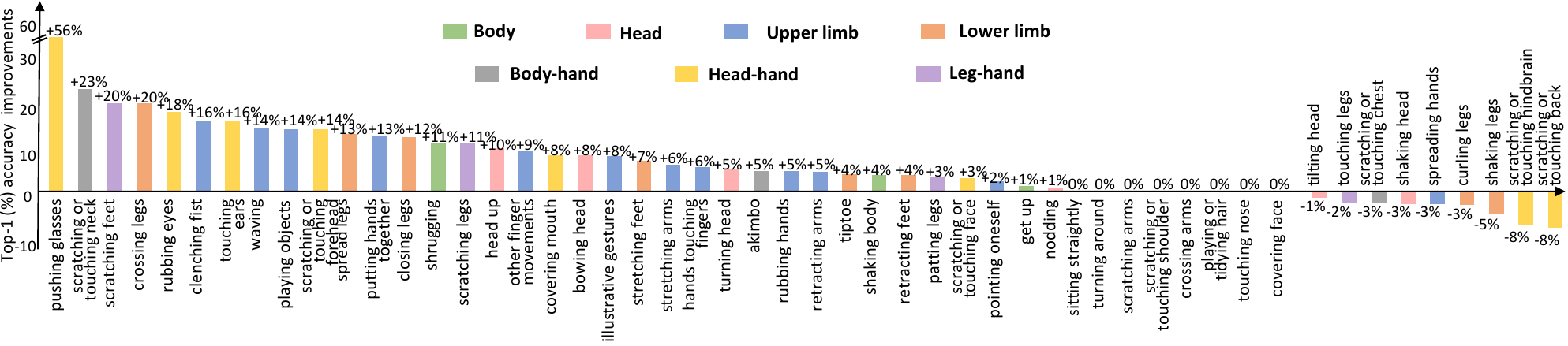}
\caption{The group-wise Top-1 accuracy improvement (\%) of our method compared to PoseConv3D on the MA-52 dataset.}
\label{fig:difference_histogram}
\end{figure*}

\subsection{Ambiguous Micro-actions Statistic Analysis}
To evaluate the effectiveness of PCAN on ambiguous micro-actions recognition, we split the action categories in MA-52 into 3 distinct difficulty levels, \ie, ``easy'', ``medium'', and ``hard''. 
Concretely, based on the recognition results of PoseConv3D~\cite{duan2022revisiting}, we gather action with accuracy lower than 50\% as the hard level, between 50\% and 60\% as the medium level, and over 60\% as the easy level. 
The comparison results are shown in Fig.~\ref{fig:easy_hard}. 
Due to the high similarity among ambiguous micro-actions, many micro-action categories are usually hard to recognize~\cite{guo2024benchmarking}. 
Compared to baseline, our method shows a significant improvement of 11.8\% in the hard-level categories, demonstrating its ability to distinguish ambiguous samples. Additionally, our method also exhibits improvements of 5.8\% and 0.6\% on the medium- and easy-level categories, indicating high recognition accuracy.

\subsection{Qualitative Analysis}

\noindent\textbf{Visualization of Recognition Results.}
As shown in Fig~\ref{fig:app_vis3}, we visualize the predicted distributions of four ambiguous categories.
In case (a), for the micro-action of ``playing object'', the focus should be on hand movements, whereas the baseline method places excessive emphasis on the associated head movement of ``bowing head''. 
Cases (b) and (c) are two ambiguous categories with extremely similar appearances. 
In contrast, our \M{} method of utilizing body- and action-level cascade prototypes can effectively constrain and calibrate the ambiguous samples, prompting the model to distinguish similar samples more effectively. Case (d) shows that although our method failed to classify this particular instance, it still achieves great improvement in all samples. We attribute this misclassification to the high similarity in motion patterns between these two samples. 
Effectively identifying and calibrating ambiguous samples in micro-action recognition tasks still remains a substantial challenge. 

\noindent\textbf{Visualization of Accuracy Distribution.}
To evaluate the recognition accuracy of \M{} on action-level categories, we also give detailed statistics of the top-1 accuracy improvement of \M{} compared to PoseConv3D. As shown in Fig.~\ref{fig:difference_histogram}, there is an improvement of 35 categories in action-level, with the same accuracy in 8 categories and a slight decrease in 9 categories.  
All body-level categories in the MA-52 dataset are boosted in subcategories. Specifically, 75\% action-level categories are boost under ``Body'' label, 66.7\% in ``Head'', 84.6\% in ``Upper limb'', 60\% in ``Head-hand'' and 75\% in ``Leg-hand''. It is obvious that \M{} improves by more than 10\% on 16 action-level categories compared to PoseConv3D, with the highest improvement of 56\% on the ``pushing glasses'' category.

\section{Conclusion}
\label{sec:conclusion}
In this paper, we presented a prototypical calibrating ambiguous network that discovers and calibrates ambiguous samples for MAR via the body- and action-level prototypes. 
The ambiguous samples are categorized into false negatives ($\mathbb{FN}$) and false positives ($\mathbb{FP}$), and we apply a hierarchical contrastive refinement to calibrate these ambiguous samples by adjusting their proximity to prototypes. 
We also presented to amplify the diversity between prototypes to enhance the model's discriminativity. 
Experimental results on the benchmark dataset showed our method corrects a significant number of hard ambiguous samples, underscoring the accuracy gains from prototype calibration. 
We believe that this work can inspire further research in other recognition tasks with a multitude of intricate samples.

\section*{Acknowledgments}
This work was supported by the National Natural Science Foundation of China (62272144, 72188101, 62020106007, U20A20183, and 62472381), the Major Project of Anhui Province (202203a05020011), the Fundamental Research Funds for the Zhejiang Provincial Universities (226-2024-00208), and the Earth System Big Data Platform of the School of Earth Sciences, Zhejiang University.

\bibliography{aaai25}

\end{document}

%% file: math_commands.tex
\usepackage{amsmath,amsfonts,bm}

\def\eqref#1{equation~\ref{#1}}

\def\1{\bm{1}}

\def\mP{{\bm{P}}}

\def\mW{{\bm{W}}}

\DeclareMathAlphabet{\mathsfit}{\encodingdefault}{\sfdefault}{m}{sl}
\SetMathAlphabet{\mathsfit}{bold}{\encodingdefault}{\sfdefault}{bx}{n}

\def\gY{{\mathcal{Y}}}

\newcommand{\R}{\mathbb{R}}

